\documentclass{article}

\usepackage{arxiv}
\usepackage[utf8]{inputenc} % allow utf-8 input
\usepackage[T1]{fontenc}    % use 8-bit T1 fonts
\usepackage{hyperref}       % hyperlinks
\usepackage{url}            % simple URL typesetting
\usepackage{booktabs}       % professional-quality tables
\usepackage{amsfonts}       % blackboard math symbols
\usepackage{nicefrac}       % compact symbols for 1/2, etc.
\usepackage{microtype}      % microtypography
\usepackage{graphicx}
\usepackage[numbers]{natbib} % Опция numbers для числового стиля цитат, eg [1]
\usepackage{doi}

\title{TRAINING A NEURAL NETWORK FOR PARTIALLY OCCLUDED ROAD SIGN IDENTIFICATION IN THE CONTEXT OF AUTONOMOUS VEHICLES}

\date{} 

\author{
  \href{https://orcid.org/0009-0001-9104-0918}{\includegraphics[scale=0.06]{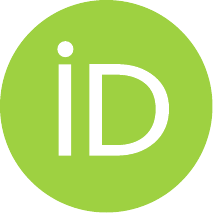}\hspace{1mm}Gulnaz Gimaletdinova} \\
  Department of Applied Mathematics and Informatics \\
  Ala-Too International University \\
  Bishkek, Kyrgyzstan \\
  \texttt{gulnaz.gimaletdinova@alatoo.edu.kg} \\
  \And
  \href{https://orcid.org/0009-0006-3534-960X}{\includegraphics[scale=0.06]{orcid.pdf}\hspace{1mm}Dim Shaiakhmetov} \\
  Department of Computer Science \\
  Ala-Too International University \\
  Bishkek, Kyrgyzstan \\
  \texttt{dim.shaiahmetov@alatoo.edu.kg} \\
  \And
  \href{https://orcid.org/0009-0007-4453-9525}{\hspace*{6.5em}\includegraphics[scale=0.06]{orcid.pdf}\hspace{1mm}Madina Akpaeva} \\
  \hspace*{5.5em}Department of Computer Science \\
  \hspace*{5.5em}Ala-Too International University \\
  \hspace*{5.5em} Bishkek, Kyrgyzstan \\
  \hspace*{5.5em} \texttt{madina.akpaeva@alatoo.edu.kg} \\ % Hypothetical correction
  \And
  \href{https://orcid.org/0009-0007-4958-1693}{\hspace*{4em}\includegraphics[scale=0.06]{orcid.pdf}\hspace{1mm}Mukhammadmuso Abduzhabbarov} \\
  \hspace*{4em}Department of Computing \\
  \hspace*{4em}Westminster International University in Tashkent \\
  \hspace*{4em}Tashkent, Uzbekistan \\
  \hspace*{4em}\texttt{mabduzhabbarov@wiut.uz} \\
  \And
  \href{https://orcid.org/0009-0002-9275-512X}{\hspace*{1em}\includegraphics[scale=0.06]{orcid.pdf}\hspace{1mm}Kadyrmamat Momunov} \\
  Department of Computer Science \\
  Ala-Too International University \\
  Bishkek, Kyrgyzstan \\
  \texttt{kadyrmamatmomunov@gmail.com} \\
}

\begin{document}

\begin{titlepage}
    \centering
    \vspace*{4cm}
    % {\large This work has been submitted to the IEEE for possible publication.\\
    % Copyright may be transferred without notice, after which this version may no longer be accessible.}
   
    {\large © 2025 IEEE.  Personal use of this material is permitted.  \\ 
    Permission from IEEE must be obtained for all other uses, in any current or future media, including reprinting/republishing this material for advertising or promotional purposes, creating new collective works, for resale or redistribution to servers or lists, or reuse of any copyrighted component of this work in other works.}

    \vspace{2em}

    {\large Related DOI: \href{https://doi.org/10.1109/CompSysTech65493.2025.11137116}{https://doi.org/10.1109/CompSysTech65493.2025.11137116}}

    \vfill
\end{titlepage}

\maketitle

\begin{abstract}
The increasing number of autonomous vehicles and the rapid development of computer vision technologies underscore the particular importance of conducting research on the accuracy of traffic sign recognition. Numerous studies in this field have already achieved significant results, demonstrating high effectiveness in addressing traffic sign recognition tasks. However, the task becomes considerably more complex when a sign is partially obscured by surrounding objects, such as tree branches, billboards, or other elements of the urban environment. In our study, we investigated how partial occlusion of traffic signs affects their recognition. For this purpose, we collected a dataset comprising 5,746 images, including both fully visible and partially occluded signs, and made it publicly available. Using this dataset, we compared the performance of our custom convolutional neural network (CNN), which achieved 96\% accuracy, with models trained using transfer learning. The best result was obtained by VGG16 with full layer unfreezing, reaching 99\% accuracy. Additional experiments revealed that models trained solely on fully visible signs lose effectiveness when recognizing occluded signs. This highlights the critical importance of incorporating real-world data with partial occlusion into training sets to ensure robust model performance in complex practical scenarios and to enhance the safety of autonomous driving.
\end{abstract}

\keywords{Road sign recognition \and partial occlusion of signs \and autonomous transportation systems \and convolutional neural networks (CNN) \and transfer learning \and road sign dataset.}

\section{Introduction}
The recognition of traffic signs directly impacts the safety and efficiency of future transportation. This field of science and technology is nearing practical application, necessitating high reliability and precision. The ability to accurately and rapidly identify traffic signs is becoming critically important for accident prevention and ensuring safe traffic flow. Computer vision technologies, such as automatic sign recognition, help mitigate the influence of human factors and enhance the interaction between vehicles and their surroundings, thereby improving the efficiency of transportation systems and road safety.

One of the primary challenges in the field of traffic sign recognition is their partial occlusion, caused by natural or anthropogenic factors. Such situations can significantly reduce the accuracy of algorithms, which is particularly critical for autonomous systems where recognition errors may lead to serious consequences. Addressing this challenge requires not only high-quality data but also advanced machine learning techniques capable of performing effectively under complex conditions. In this article, we focus on the development and evaluation of models designed to handle the recognition of both fully visible and partially occluded traffic signs.

To address the stated problem, we collected and made publicly available a dataset \cite{use_analysis} that includes both fully visible and occluded traffic signs. Based on this dataset, we trained our custom neural network as well as several pretrained models using transfer learning, including YOLO11x, EfficientNetB0, ResNet, and others.

In addition to this, we addressed the following research questions: 

\begin{itemize}
  \item RQ1: Can a neural network trained solely on fully visible traffic signs successfully recognize partially occluded traffic signs?
  \item RQ2: Is a neural network trained on synthetically generated images simulating occluded traffic signs capable of effectively recognizing traffic signs with partial occlusion in real-world conditions? 
\end{itemize}

\section{Related Work}
This section presents a concise summary of previous research on traffic sign detection. Given the critical role of traffic signs in ensuring road safety, numerous studies have focused on their accurate and efficient recognition.

\subsection{Deep Learning Approaches for Fully Visible Traffic Signs}

A considerable amount of research has already been conducted with excellent results. \cite{gong2022traffic} proposed a pyramid pooling structure that is integrated into the YOLOv3 network to combine local and global features effectively. The improved algorithm is compared with YOLOv3 and other popular object detection methods. Results show that the enhanced YOLOv3 achieves a mean average precision (mAP) of 77.3\%, which is 8.4\% higher than the original YOLOv3. Another study introduced a lightweight traffic sign recognition model with an architecture named Dynamic Feature Extraction-Efficient Vision Transformer (DFE-EViT), which consists of two components: a dynamic feature extraction network and an Efficient Vision Transformer (EViT) classifier \cite{ge2023lightweight}. This architecture is designed to solve complex and dynamic traffic sign recognition tasks, by combining local details with global receptive fields. The model performs well, achieving 98.4\% accuracy with only 0.859M parameters.

Traditional models for traffic sign detection and recognition rely on deep neural networks. For example, \cite{megalingam2023indian} implemented an end-to-end learning approach using a Refined Mask R-CNN (RM R-CNN) based on Convolutional Neural Networks (CNN) which achieved an accuracy of 97\%. Another example of using CNN can be the LeNet-5 network model used to recognize traffic signs \cite{an2024lightweight}. The model is tested on the German Traffic Sign Recognition Benchmark (GTSRB) database, achieving a recognition accuracy of 97.53%.

Karthika and Parameswaran \cite{karthika2022novel} suggested a novel convolutional neural network model. This study is carried out in two stages: sign detection and sign recognition. In the detection phase, the YOLOv3 architecture is employed to identify traffic signs. The output from this stage is then fed into a CNN for the recognition phase. The model achieved an accuracy of 86.6\%.

A large number of studies have been conducted using the GTSRB or Belgium Traffic Sign Classification (BTSC) database for training models.
For example, in one of the experiments, Real-Time Image Enhanced CNN (RIECNN) is utilized for Traffic Sign Recognition \cite{abdel2022riecnn}. Experiments are performed using datasets such as the GTSRB, the BTSC, and the Croatian Traffic Sign (rMASTIF) benchmark, achieving recognition accuracies of 99.75\% for GTSRB, 99.25\% for BTSC, and 99.55\% for rMASTIF. 

\subsection{Recognizing Traffic Signs Under Partial Occlusion}

Traffic sign detection and recognition tasks become challenging when the target sign is partially blocked by nearby objects such as trees, structures or obstacles. The following section presents a review of studies in which models have been trained specifically on partially occluded signs.
In one of the studies, generative models were combined with deep neural networks (DNN) \cite{sanyal2024indian}. Specifically, the final convolution layers of the DNN were replaced with a differentiable compositional model. This integrated approach focuses on recognizing the uncovered parts of traffic signs. The proposed model, called Generative DNN (GDNN), performs better than traditional DNNs when dealing with images of partially covered traffic signs. The GDNN's effectiveness was also compared with other methods to show its superior performance. 

Another work presents traffic and road sign recognition using the LeNet network \cite{fredj2023efficient}. A novel dataset, called the Tunisian Traffic Signs Dataset, is used to evaluate the model's performance. Additionally, the following conditions were taken into account: diverse weather conditions, intricate backgrounds, fluctuating illumination, fading sign colors, occlusion and missing parts.

Another study introduces an automated method that prioritizes visible parts while disregarding obstructed portions of partially occluded traffic signs. \cite{mannan2022cognition}. Features are extracted using a convolutional neural network-based invariant feature extraction technique. The researchers have created two datasets by combining originally and synthetically occluded images collected from field surveys and the well-known GTSRB database. For both visible and occluded signs, the method achieves an average precision of 0.81 and a recall of 0.79.

\cite{luo2023detection} presented an algorithm for extracting obscured traffic sign information from road images and integrating it into the vehicle movement process using speed data. The proposed approach identifies the region of interest (ROI) through a color-shape-based traffic sign detection and recognition algorithm. Multi-frame road images are then merged to reconstruct a complete traffic sign image. The findings indicate that the proposed algorithm effectively restores obscured traffic signs in motion while preserving their geometric characteristics.

\section{Methodology}
In this chapter, we describe the dataset \cite{use_analysis} we collected, which consists of 5,746 photographs, and address the topics of utilizing pretrained models and developing our own architecture.

\subsection{Dataset}

For the study of recognizing partially occluded traffic signs, we collected the dataset consisting of images of both fully visible and partially occluded traffic signs. An example of photographs from the dataset is shown in Figure 1.

\begin{figure}[h]
    \centering
    \includegraphics[width=0.7
    \textwidth]{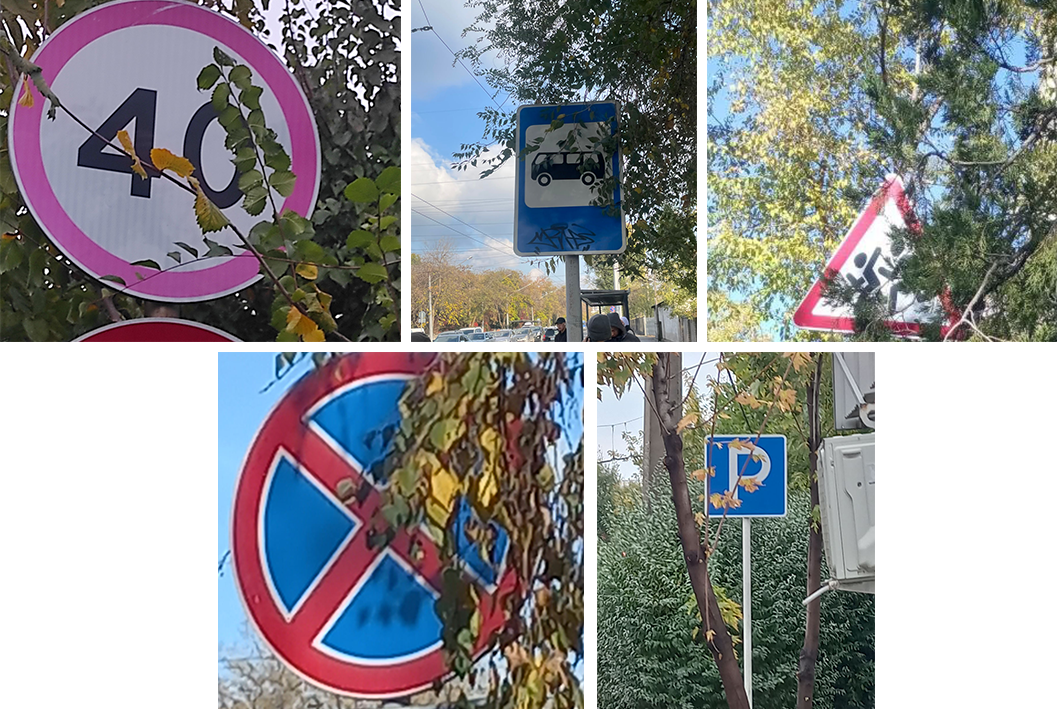} % Замените на имя вашего файла
    \caption{Examples of images from the dataset.}
    \label{fig:example}
\end{figure}

To collect the data, volunteers were enlisted to photograph both fully visible and occluded traffic signs in real urban conditions. A total of 5,746 images of 26 different traffic signs were gathered, with maximum resolution images approximately 3,472 × 4,640 pixels and minimum resolution images around 475 × 442 pixels.

The dataset was manually labeled and sorted into classes. The class with the largest number of fully visible traffic signs contains 892 images, while the class with the smallest number has 1. For occluded signs, the class with the largest number includes 1,243 images, and the class with the smallest number has 7.

The collected dataset is publicly available and can be downloaded at the following link: \href{https://www.kaggle.com/datasets/madinaakpaeva/traffic-sign-images/}{https://www.kaggle.com/datasets/madinaakpaeva/traffic-sign-images/}.

\subsection{Data Augmentation }
To enhance the diversity of the data, augmentation was employed. Data augmentation is the process of artificially expanding the volume of training data and improving its variety by applying various transformations. These transformations may include changes in geometry, color, contrast, or scale of the images. The primary objective of augmentation is to generate additional variations of the original data, enabling the model to learn from a broader range of examples, which aids in generalizing knowledge and preventing overfitting.\cite{mumuni2022data}. In our study, the following augmentations were utilized: Random Translation, Random Contrast, and Random Zoom.

\subsection{Training Methods and Development of the Optimal Model}

Working with a specific set of 26 types of traffic signs requires a meticulous approach to dataset formation and its optimization for AI training. The choice between adapting existing models and developing a custom architecture determines the subsequent direction of the research.

\subsubsection{Preparation and Optimization of the Road Sign Dataset}

The original dataset comprises 26 types of traffic signs, categorized into fully visible and occluded. However, their distribution across classes reveals a clear imbalance. The three largest classes contain 2,960 images, accounting for 51.5\% of the total volume. In contrast, the ten smallest classes consist of 251 images, representing 4.4\% of the total number of photographs. This distribution indicates a pronounced imbalance in the dataset.

To avoid working with imbalanced data and to optimize the training process, 12 types of signs that are sufficiently represented in our dataset were selected for further study. These signs include Bus Stop, Children, Give Way, Turn Left, Main Road, Speed Limit 40, No Stopping, Parking Lot, Pedestrian Crossing, Turn Right, Speed Hump, and Straight. Additionally, signs from our dataset were supplemented with signs from an open dataset \cite{road_sign_dataset}.

\subsubsection{Architecture of the Custom Neural Network}

Initially, we focused on developing our own neural network. Its architecture is presented in Figure 2.

\begin{figure}[h]
    \centering
    \includegraphics[width=1\textwidth]{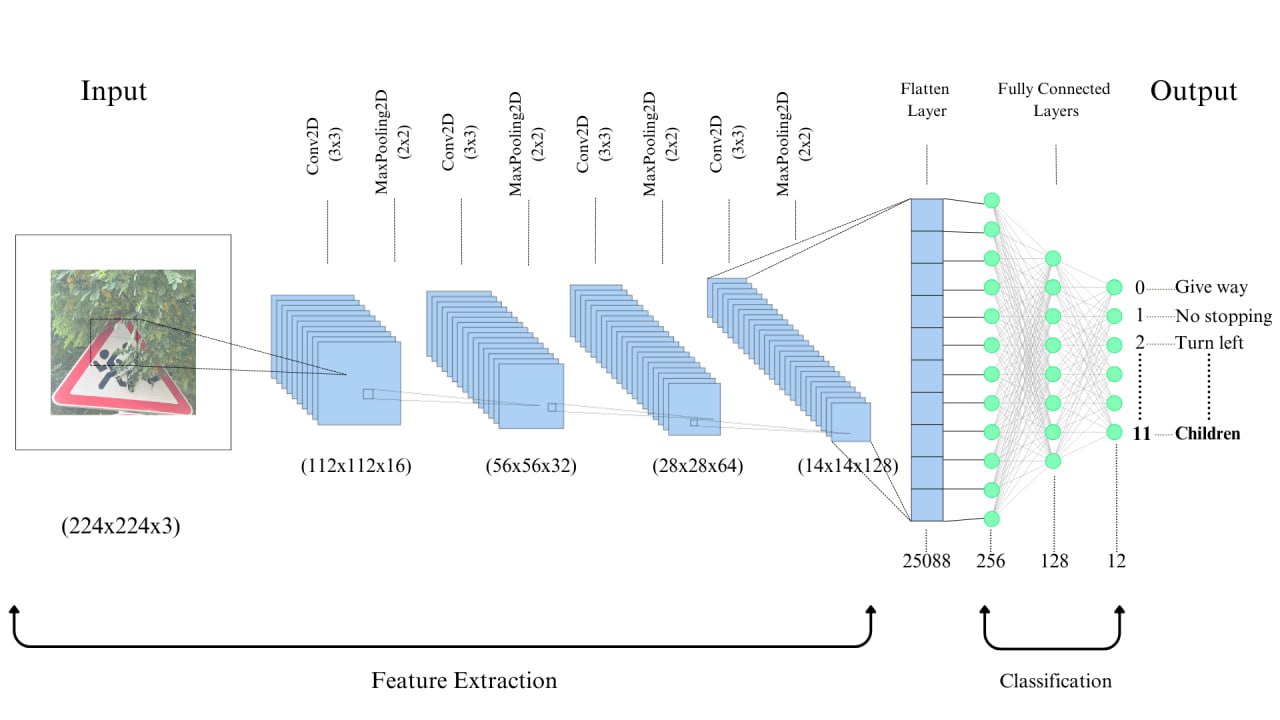} % Замените на имя вашего файла
    \caption{Architecture of the proposed CNN model for road sign recognition.}
    \label{fig:example}
\end{figure}

To develop this architecture, tests were conducted on various combinations of hyperparameters, followed by their tuning using Grid Search. The best results were achieved by a model constructed with four convolutional layers and two hidden layers using the ReLU activation function, employing the CrossEntropyLoss function and the Adam optimization method. The trained model's accuracy reached 96\%.

\subsubsection{Transfer Learning}

Transfer Learning is an approach in machine learning where a model trained on one dataset for a specific task is repurposed to address a different but related task. In this context, the model transfers previously acquired knowledge and features extracted from the original task to the new task \cite{zhu2023transfer}. This method significantly reduces training time and enhances the model's efficiency.

As part of the study, we tested various approaches to address the set tasks, including the use of pretrained models. In these models, the output layer was unfrozen and replaced, followed by additional training on our dataset. The results obtained were unsatisfactory: transfer learning yielded an accuracy of 50.63\% for EfficientNetB0 and 51.9\% for ResNet.

Based on the obtained data, it was decided to retrain the model using transfer learning, but with all layers unfrozen. Various models were tested with all layers opened for fine-tuning. The best results were demonstrated by the EfficientNetB0, VGG16, and MobileNet models. The highest accuracy was achieved by the VGG16 model, which attained a result of 99\% without signs of overfitting.

\section{Results}

We collected and made publicly available the dataset \cite{use_analysis} consisting of 5,746 images of 26 different fully visible and occluded traffic signs. 

The CNN model architecture we proposed demonstrated high accuracy in recognizing partially occluded traffic signs, achieving an accuracy score of 96\%. However, the VGG16 model, trained using transfer learning with full layer unfreezing, surpassed this result, reaching an accuracy of 99\%. This makes it the most effective solution for recognizing partially occluded traffic signs. 

In addition to developing the neural network, we also conducted research to address the research questions outlined in this article. The experiments carried out enabled us to obtain answers to the posed questions.

RQ1: Can a neural network trained solely on fully visible traffic signs successfully recognize partially occluded traffic signs?  

To address RQ1, we trained several models using a dataset consisting solely of fully visible traffic signs. 

The YOLO11x model, trained on 11,000 images of fully visible traffic signs, exhibited poor performance in recognizing partially occluded signs, particularly when the occlusion level exceeded 50-60\% (Figure 3). Its accuracy was 64\% for occluded signs, whereas it achieved an accuracy of 95\% for fully visible signs.

\begin{figure}[h]
    \centering
    \includegraphics[width=1\textwidth]{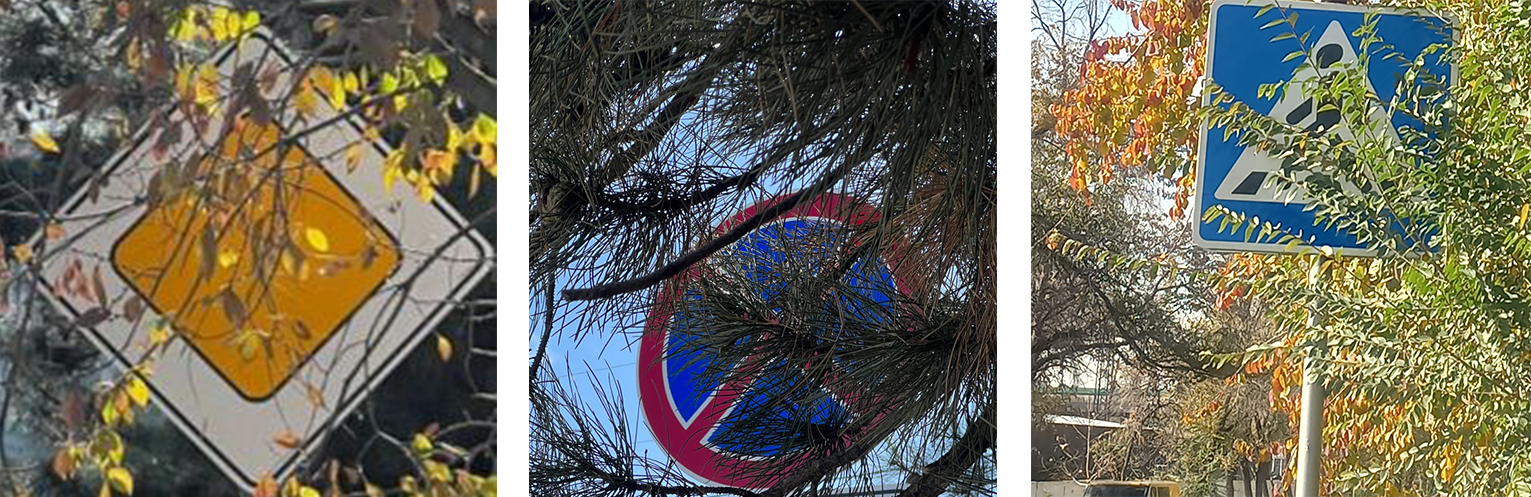} % Замените на имя вашего файла
    \caption{Examples of images that the model failed to recognize correctly.}
    \label{fig:example}
\end{figure}

The EfficientNetB0 and ResNet models, fine-tuned on fully visible traffic signs, also failed to provide the necessary accuracy when recognizing occluded signs. The accuracy of these models was 56.25\% and 52.08\%, respectively.

Thus, based on the experimental results, it can be concluded that for successful recognition of occluded signs in real-world conditions, it is essential to utilize a dataset that includes both fully visible and occluded traffic signs.

RQ2: Is a neural network trained on synthetically generated images simulating occluded traffic signs capable of effectively recognizing traffic signs with partial occlusion in real-world conditions?  

To test the hypothesis of RQ2, Python code was developed to overlay various objects (squares, stripes, etc.) onto images of fully visible traffic signs, simulating their partial occlusion. However, the results indicated that models trained on such data did not achieve high recognition accuracy. The YOLO11x model and our custom neural network, trained on simulated data, exhibited low accuracy when tested on real images of occluded signs, with results ranging from 55\% to 82\%. This outcome demonstrates that relying solely on simulated data is not an effective solution to the problem of insufficient photographs of partially occluded traffic signs.

Thus, the experiments conducted for RQ1 and RQ2 revealed that for effective recognition of occluded traffic signs, the dataset must include real images of occluded signs.

\section{Conclusion}
As part of our research, we collected and made publicly available a dataset \cite{gong2022traffic} containing images of traffic signs, including those partially occluded by elements of urban infrastructure. This represents a significant contribution, as in real-world conditions, traffic signs are frequently obscured by trees, billboards, or other objects, while existing datasets rarely account for such scenarios. Our dataset combines both fully visible and partially occluded signs, making it more versatile and valuable for training models capable of operating in complex environments.

Based on this dataset, we conducted several experiments in model training. The VGG16 model, fine-tuned with full unfreezing of all layers, demonstrated the highest accuracy—99\%—without signs of overfitting.

We also addressed the key questions RQ1 and RQ2, establishing that it is critically important to use datasets containing both fully visible and partially occluded signs for training models. Without this, autonomous driving algorithms may fail in real-world conditions where sign visibility can be limited.

The obtained results provide a solid foundation for future research and the development of traffic sign recognition methods in challenging conditions, bringing us closer to the creation of reliable autonomous transportation systems.

\section{Future Work}

It is necessary to collect more occluded traffic signs for the open-source dataset and to continue experimenting with occlusion simulation, as we do not rule out the possibility that effective methods for this can be identified.

Our dataset requires additional photographs to achieve data balance (Figure 2). To expand the dataset, both fully visible and partially occluded images of traffic signs are needed. This will enable the model to better distinguish signs under various conditions and improve the accuracy of their recognition.

Our research has established a robust foundation for recognizing partially occluded traffic signs; however, several aspects require further refinement. Firstly, expanding the dataset to include various weather conditions (rain, fog, snow) and lighting levels (nighttime, twilight) would enable an assessment of model resilience to additional factors affecting sign visibility in real-world scenarios.

Secondly, it would be beneficial to explore the effectiveness of hybrid approaches that combine synthetic and real data. Although the RQ2 experiments demonstrated the limited efficacy of fully synthetic data, integrating synthetic augmentation with realistic textures and urban environment objects could offer a solution for increasing the training sample size without the need to collect a large number of new photographs.

Furthermore, future research could focus on testing advanced architectures, such as YOLOv11x with enhanced attention mechanisms or transformer-based models, to compare their performance with the results achieved. This would help determine whether new approaches can surpass the current accuracy or enhance the generalization capability of the models.

Finally, the integration of contextual information, such as data on the position of signs relative to the road or surrounding objects, could be considered to improve sign recognition under significant occlusion. These research directions have the potential to elevate traffic sign recognition technologies to a new level, making them more applicable to the challenges of autonomous transportation.

\bibliographystyle{unsrtnat}
\bibliography{references}

\end{document}